\documentclass[preprint,12pt]{elsarticle}

\usepackage[margin=1in]{geometry}
\usepackage{graphicx}
\usepackage{booktabs}
\usepackage{array}
\usepackage{longtable}
\usepackage{amsmath}
\usepackage{microtype}
\usepackage{textcomp}
\usepackage{xurl}
\usepackage[colorlinks=true,linkcolor=blue,citecolor=blue,urlcolor=blue]{hyperref}
\newcolumntype{P}[1]{>{\raggedright\arraybackslash}p{#1}}
\newcommand{\cotwo}{CO\textsubscript{2}}

\journal{Smart Agricultural Technology}
\biboptions{numbers,sort&compress}

\begin{document}

\begin{frontmatter}

\title{Calibration-First Reward-Component Auditing for Reinforcement Learning Control in Smart Greenhouses}

\author[cuaf]{Yuhui Bie}
\author[cuaf]{Guowei Xu}
\author[cuaf]{Yaojun Wang\corref{cor1}}
\cortext[cor1]{Corresponding author.}
\ead{wangyaojun@cau.edu.cn}

\address[cuaf]{College of Information and Electrical Engineering, China Agricultural University, Beijing, China}

\begin{abstract}
Greenhouse reinforcement learning can test climate-control ideas at a speed and scale that is difficult to achieve with crop experiments alone. For smart-greenhouse control, however, a single simulator return is not enough: a grower or control engineer also needs to know when the policy heats, enriches CO\textsubscript{2}, vents, manages humidity, deploys screens, or uses lamps.

We propose a reproducible calibration-first reward audit framework that keeps named greenhouse-control reward components comparable across simulator training, facility-adapted rollouts, logged Autonomous Greenhouse Challenge records, and actuator-rule distillation. In GreenLight-Gym, the framework decomposes the scalar reward into conditional temperature, CO\textsubscript{2}, humidity and vapor-pressure-deficit, screen, and actuation-proxy terms; adapts GreenLight to the second Autonomous Greenhouse Challenge logged climate traces; and scores the same components on logged greenhouse data.

The evidence comes from three validation layers. Component rewards did not break PPO learnability: nine 500k-step runs remained above the rule-controller floor. In AGC2 blocked validation, the facility-adapted GreenLight candidate improved state-rollout accuracy across four unique team/window traces represented by seven preprocessing validation contexts. Mean root-mean-square error decreased by $0.184^\circ$C for temperature, 4.3~pp for relative humidity, and 563~ppm for CO\textsubscript{2}, with no context-level temperature degradation. Reward-component distances to real-log values decreased at both timestep and daily-sum scales. The clearest simulator-side behavioral result came from shallow-tree actuator distillation: thermal-screen, lamp, and blackout-screen actions were more rule-approximable under the physics-prior reward, with $R^2$ increasing from 0.652 to 0.835, from 0.631 to 0.838, and from 0.592 to 0.750.

Logged-data scoring further localized crop- and channel-specific alignment signals, including a positive but underpowered AGC1 cucumber rank correlation, $\rho=0.829$ with exact permutation $p=0.058$, and calibration targets in AGC2 and AGC4. Together, these results support reward-component auditing as a pre-deployment diagnostic layer for greenhouse RL, not as evidence of field-level yield improvement. The framework reports how a learned policy earns reward, where simulator calibration changes the conclusion, and which crop or actuator channels should be checked before facility-specific controller trials.
\end{abstract}


\end{frontmatter}

\section{Introduction}
\label{sec:introduction}

Greenhouse climate control is more than keeping one variable near one setpoint. Heating, ventilation, CO\textsubscript{2} dosing, screens, lighting, crop response, and operating cost all act through the same climate system. Classical optimal-control and model-based-control studies make this coupling explicit by specifying the greenhouse and crop model before optimizing operation \citep{Bakker1995GreenhouseClimate,VanStraten2010OptimalControl,Jones1991TOMGRO,Vanthoor2011GreenhouseClimateModel}. Energy-aware control depends on the same modelling choices: conclusions change with the representation of heat exchange, humidity, CO\textsubscript{2}, crop growth, and weather forcing \citep{Iddio2020GreenhouseEnergyReview}. Reinforcement learning (RL) is attractive because simulation can generate many policy rollouts without putting a crop at experimental risk. Recent greenhouse studies have therefore compared RL with model predictive control (MPC), combined RL and MPC, tested robust or model-based variants, and released benchmark environments for controller development \citep{An2021SimulatorPlanning,Zhang2021RobustMBRL,Morcego2023RLMPC,Ajagekar2023RobustDRL,Mallick2024RLMPC,Mansour2025AdaptiveRobustGreenhouse,GreenLightGym}.

This coupling also makes a high simulator return difficult to interpret. A grower or control engineer does not only ask whether the return increased; they ask what the controller did to obtain it. Heating, screen closure, CO\textsubscript{2} enrichment, ventilation, humidity control, and supplemental lighting have different physical meanings and different failure modes. CO\textsubscript{2} enrichment, for example, depends on light and crop state, and high light under unsuitable temperature can create stress rather than useful assimilation \citep{Mortensen1987CO2,Pan2019CO2LightTomato,Niu2023LightSpectraTomato,Lu2017PhotoinhibitionTomato}. Once these effects are collapsed into a single profit-like reward, a small return gain may reflect a physiological response, a shifted setpoint, a screen-use shortcut, a favorable weather episode, or a scaling artifact. This paper therefore asks a practical reporting question: can the reward terms used for simulator training remain interpretable and comparable when they are checked against calibrated simulator rollouts and logged greenhouse records?

Process-based greenhouse simulators such as GreenLight \citep{Katzin2020GreenLight} provide a useful starting point because they expose climate states and actuator channels. GreenLight-Gym \citep{GreenLightGym} wraps GreenLight as an OpenAI Gym environment with six controls: heating, CO\textsubscript{2}, thermal screen, ventilation, lighting, and blackout screen. Its default reward is still a scalar that combines growth revenue and energy cost, so it does not show which physiological or actuator terms drive the policy. Reward shaping can make learning easier, but it can also change what the agent optimizes unless the shaping assumptions remain visible \citep{Ng1999RewardShaping,Wiewiora2003Potential,Devlin2012DynamicPotential,Schulman2017PPO}.

\subsection{Related work and gap}

Greenhouse RL has begun to move from proof-of-concept policy learning toward benchmarked comparisons with MPC and robust optimization. GreenLight-Gym provides fixed weather inputs, an actuator interface, and a rule-controller baseline for simulator studies \citep{GreenLightGym}. Recent work also makes clear that nominal return is not enough. A controller must respect climate constraints, remain usable under model mismatch, and be interpretable enough for growers to inspect or adjust \citep{Morcego2023RLMPC,Ajagekar2023RobustDRL,Xiao2026GrowerLoop}. These expectations are familiar in model-predictive control, where constraints and forecasts are explicit, and in safe RL, where deployment limits are part of the learning problem \citep{VanStraten2010OptimalControl,Garcia2015SafeRL}.

Two prior lines of work motivate treating reward design as an audit target. The first is sim-to-real caution. Robotics and control studies often use simulation for safe data generation, but policies can fail when simulator dynamics, actuation, sensing, or aggregation differ from the target system \citep{Tobin2017DomainRandomization,Salvato2021RealityGap,Muratore2022RandomizedSimReview}. The same issue appears in greenhouse simulation: a reward component computed every 15 minutes in simulation cannot be assumed comparable with a daily aggregate score from logged greenhouse data. The second line is explainable RL. Policy extraction, decision-tree distillation, and broader XRL taxonomies provide tools for inspecting learned behavior without treating the neural policy itself as transparent \citep{Bastani2018VIPER,Puiutta2020XRLSurvey,Bekkemoen2024XRL}. In agricultural modelling, explainable AI is also used to show which measured variables support a prediction rather than treating black-box accuracy as sufficient \citep{Ryo2022AgXAI}. In greenhouse control, this leads to a channel-wise question: can the learned screen, lamp, CO\textsubscript{2}, or ventilation action be approximated by a shallow rule under the same simulator distribution?

Reward shaping can encode domain knowledge, but the shaping assumptions also need to remain inspectable. Potential-based shaping preserves policy invariance under clean conditions \citep{Ng1999RewardShaping,Devlin2012DynamicPotential}. Greenhouse reward terms rarely meet such conditions because growth, weather, climate comfort, and actuator costs are coupled. Decomposed rewards can therefore serve a diagnostic role: physiological terms can be inspected individually, and the resulting actions can be checked for shallow-rule approximability \citep{Garcia2015SafeRL,Juozapaitis2019RewardDecomposition,Lin2019RewardDecomposition}.

The missing step is a practical reward check. Existing greenhouse RL studies often report aggregate simulator return, but rarely show whether reward terms have consistent semantics across simulator rollouts and logged greenhouse records, whether an energy term is a measured quantity or an action proxy, or whether a small return margin can be traced to recognizable actuator behavior. This study uses the same named components for three connected tasks: training a GreenLight-Gym policy, scoring AGC greenhouse logs, and inspecting actuator behavior with shallow rules. The audit asks three bounded questions: whether the reward remains learnable, whether action changes occur in interpretable regimes, and whether logged data point to crop-specific calibration targets.

\subsection{Contribution and scope}

This study contributes a calibration-first reward-component audit protocol for greenhouse RL, designed to test whether a reward remains learnable, comparable across simulator and logged records, and interpretable at actuator level before controller-transfer claims are made.

\begin{figure}[htbp]
\centering
\includegraphics[width=\textwidth]{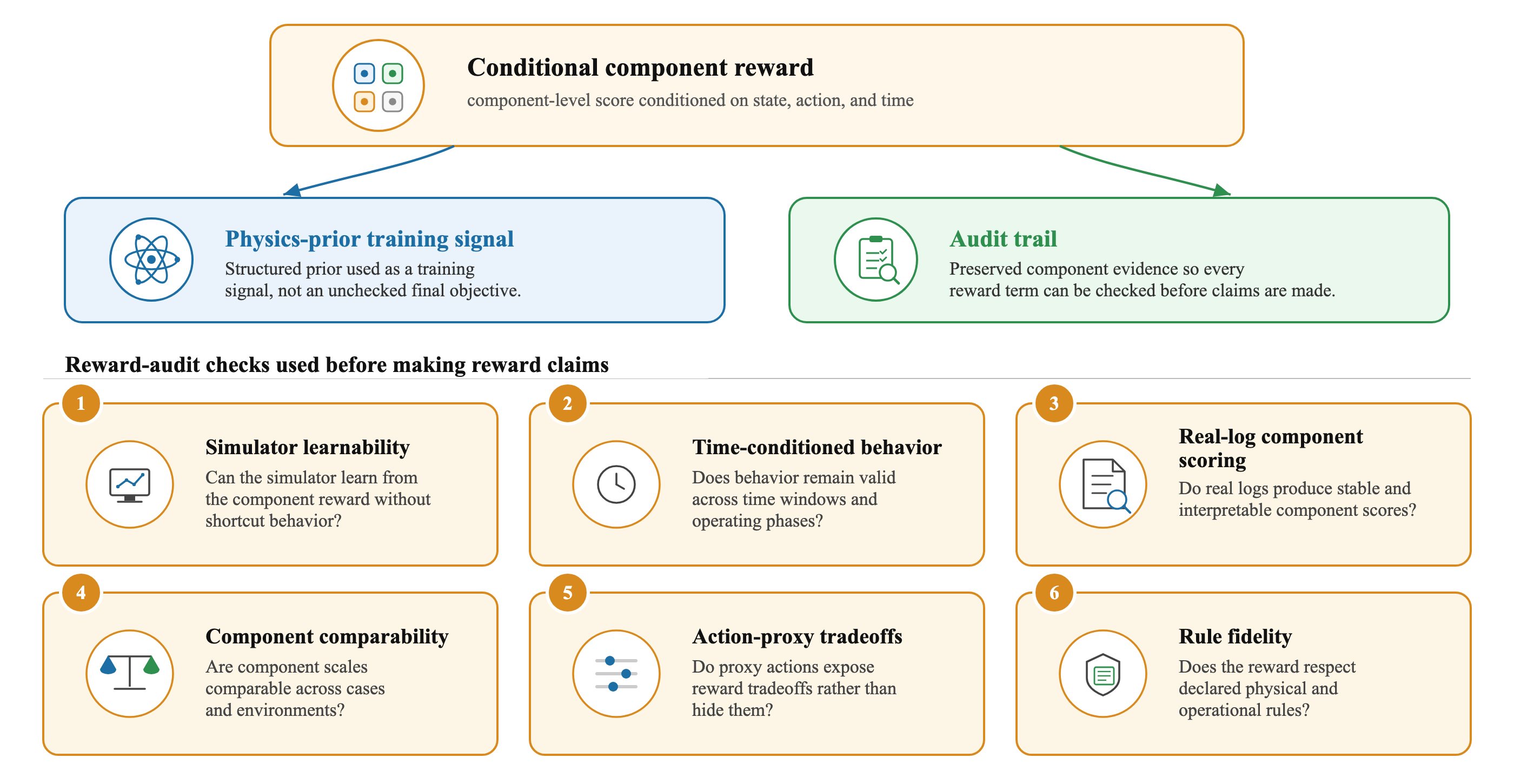}
\caption{Calibration-first reward-audit workflow used in this study. A conditional component reward is retained both as a physics-prior training signal and as a record of component scores. Six checks are reported before reward claims are made: simulator learnability, time-conditioned behavior, real-log component scoring, component comparability, action-proxy tradeoffs, and rule fidelity.}
\label{fig:diagnostic_workflow}
\end{figure}

The framework has three steps (Fig.~\ref{fig:diagnostic_workflow}). First, the scalar GreenLight-Gym objective is decomposed into named components for temperature, CO\textsubscript{2}, humidity/VPD, screen use, and weather-normalized actuation. Second, reward interpretation is gated by adapting GreenLight to AGC2 logged climate traces before simulator rollouts are used as evidence. Third, the same components are recomputed on AGC1, AGC2, and AGC4 greenhouse logs \citep{Hemming2019AGC1,Hemming2020AGC2,Maree2025AGC4}, and learned actuator channels are distilled into shallow rules so screen, lamp, heating, ventilation, and CO\textsubscript{2} behavior can be inspected directly rather than inferred from scalar return alone.

\section{Materials and methods}
\label{sec:methods}

\subsection{Simulator environment and action interface}

Experiments use GreenLight-Gym, a reinforcement-learning benchmark built around the GreenLight process model. GreenLight represents greenhouse climate, crop growth, lighting, heat exchange, and weather forcing. GreenLight-Gym exposes these dynamics through state observations, a continuous actuator vector, and the official rule-based controller distributed with the benchmark. The action vector contains six normalized controls: pipe heating, CO\textsubscript{2} dosage, thermal screen, ventilation, supplemental lighting, and blackout screen. Because these controls are normalized simulator actions, action-cost analyses are reported as actuation-proxy diagnostics rather than as physical resource accounting.

The simulator is used here as a controlled test bed for reward design and calibration analysis. Process-based models are useful because they expose state variables and actuator channels, but their numerical outputs still depend on equations, timestep resolution, weather forcing, and unit conventions. The analysis keeps three quantities separate: native simulator reward, component reward terms used for inspection, and physical greenhouse resource use.

The baseline controller is the official rule-based policy distributed with the benchmark. A learned policy must clear this floor before its component behavior is interpreted. This requirement keeps the analysis tied to a policy that is both learnable and inspectable, rather than to traces that are easy to explain but poor controllers. Evaluation summaries report native simulator reward together with component contributions, radiation-conditioned actions, and rule-fidelity scores.

\subsection{Conditional component reward}

The training reward keeps the native GreenLight-Gym reward visible while adding a diagnostic physics-prior term:
\begin{equation}
r_t = 0.7 r_t^{\mathrm{native}} + 0.3 r_t^{\mathrm{diag}}, \qquad
r_t^{\mathrm{diag}} = \sum_{k \in \mathcal{K}} w_k r_{t,k}.
\end{equation}
Here $t$ indexes the 15-min simulator timestep, $r_t^{\mathrm{native}}$ is the original GreenLight-Gym scalar reward, $r_t^{\mathrm{diag}}$ is the diagnostic physics-prior term, $\mathcal{K}$ is the set of named reward components, $r_{t,k}$ is the timestep score for component $k$, and $w_k$ is its component weight. The 0.7/0.3 mixture was used as a fixed audit-oriented design choice to keep the native GreenLight-Gym objective dominant while exposing diagnostic component traces; it was not tuned against AGC outcomes. The mixture is not proposed as an optimal greenhouse-control reward. Sensitivity to actuation-proxy weighting is reported in Additional file 1, and broader mixture optimization is outside the present audit claim. All component weights were fixed at $w_k=1$ unless explicitly varied in the actuation-proxy sweep. No component weight was fitted to AGC yield, AGC team ranking, or any logged greenhouse outcome.

The diagnostic term records seven components: temperature comfort, temperature direction, CO\textsubscript{2} direction, humidity/VPD, light, screen, and actuation proxy (Table~\ref{tab:component_reward_terms}). Daytime in simulator rollouts is defined as outside radiation $>50$~W~m$^{-2}$. VPD is computed as
\begin{equation}
\mathrm{VPD}(T,RH)=0.6108\exp\left(\frac{17.27T}{T+237.3}\right)(1-RH/100),
\end{equation}
with $T$ in $^\circ$C, $RH$ in percent, and VPD in kPa.

\begin{table}[htbp]
\centering
\caption{Reward-component audit map used for simulator training and AGC-log scoring. The table keeps the main text focused on component purpose and gating logic; exact timestep formulas, weights, and daily proxy formulas are listed in Additional file~1.}
\label{tab:component_reward_terms}
\small
\setlength{\tabcolsep}{4pt}
\begin{tabular}{P{0.18\linewidth}P{0.31\linewidth}P{0.25\linewidth}P{0.18\linewidth}}
\toprule
Component & Simulator-side gate & AGC-log analogue & Diagnostic use \\
\midrule
Temperature comfort & Crop-specific day/night setpoints and a radiation-derived day flag. & Daily day/night temperature summaries compared with crop setpoints. & Thermal comfort. \\
Temperature direction & Heating and ventilation receive opposite signs when temperature is outside a comfort band. & Daily heating and ventilation proxies scored under warmer/colder deviations. & Corrective heating/ventilation direction. \\
CO\textsubscript{2} direction & CO\textsubscript{2} dosing is rewarded under daylight, sufficient radiation, low ventilation, and below-target CO\textsubscript{2}. & Daily dosing and ventilation proxies evaluated under crop light thresholds. & Light-supported, low-vent enrichment. \\
Humidity/VPD & VPD comfort band and outside-band ventilation responses. & Daily mean temperature/RH converted to VPD and scored by deviation direction. & Humidity-stress diagnostic. \\
Light & Lamp use is rewarded under low daylight radiation and discouraged at night. & Daily lamp proxy scored on low-radiation days. & Lighting-use signal. \\
Screen & Day shade and night-insulation gates depend on radiation and temperature. & Thermal-screen percentage scored under high-radiation, low-radiation, and night gates. & Screen-rule behavior. \\
Actuation proxy & Weather-normalized action burden from heating, CO\textsubscript{2}, and lamp channels. & Daily heating, CO\textsubscript{2}, and lamp proxies combined with cold/dark factors. & Normalized control burden. \\
\bottomrule
\end{tabular}
\end{table}

\subsection{Reward-component implementation}

The component reward is implemented as a wrapper on top of the scalar GreenLight-Gym reward. At each timestep, component scores are computed from the simulator state, weather context, and normalized action vector. Their sum enters the training reward, while the individual component values are retained for ablation, one-out ranking, and comparability checks.

For AGC logs, the same component names are recomputed from daily climate summaries and actuator proxies. Cucumber uses $24.0/19.0^\circ$C day/night setpoints and a radiation threshold of 3 in the daily CO\textsubscript{2} gate; tomato and cherry tomato use $21.5/18.5^\circ$C and threshold 5. When direct physical flow is unavailable, logged heating and CO\textsubscript{2} values are treated as setpoint-derived proxies, allowing the same reward map to locate calibration targets in public greenhouse records.

\subsection{Training protocol}

Policies are trained with Proximal Policy Optimization (PPO) \citep{Schulman2017PPO}. The official rule-based controller provides the base action, and the neural policy learns a bounded residual ($\pm 0.15$ scale). VecNormalize is applied to observations and rewards. Training uses three start days (120, 180, 240) and three evaluation seeds (42, 123, 456). The start days evaluate the controller across different seasonal radiation and temperature regimes rather than tuning it for a single weather episode. Evaluation rollouts are summarized by native reward, action traces, and component scores.

The rule-controller floor is the official GreenLight-Gym rule-based policy executed in the same simulator, weather, start-day, and evaluation setting as the learned policies. It is retained as a sanity check: a learned policy must beat this floor before component behavior is interpreted. Native reward is used for learnability and floor-clearing, whereas component terms are used for inspection. This separation matters because component traces can be informative even when the native-reward margin is small.

\subsection{Configurations}

Three configurations are evaluated. The tomato component-reward wrapper uses a $21.5/18.5^\circ$C day/night set-point pair, the cucumber component-reward wrapper uses a $24.0/19.0^\circ$C pair, and the scalar-reward tomato baseline uses the same tomato temperature pair.

\subsection{Diagnostic analyses}

The evaluation uses six reward-audit checks instead of one performance score. First, each reward wrapper must clear the rule-controller floor after 500k training steps. Second, rollout actions are stratified by hour and solar-radiation bin to test whether the gated CO\textsubscript{2} term changes behavior in a physiologically plausible regime. Third, one-out component ablation checks whether any single component dominates the normalized reward scale; the full ranking is reported in Additional file~1, the supplementary PDF accompanying this manuscript.

The remaining checks ask whether the reward can still be read after learning and across domains. A reward--actuation-proxy sweep evaluates tradeoffs in normalized action space, where the horizontal axis is a simulator action proxy rather than measured energy. Policy distillation fits shallow decision trees to each actuator channel \citep{Bastani2018VIPER,Puiutta2020XRLSurvey,Bekkemoen2024XRL}. Finally, the same named components are scored on logged AGC greenhouse data \citep{Hemming2019AGC1,Hemming2020AGC2,Maree2025AGC4} and compared with team-level yield rankings using Spearman correlation. Before that cross-domain comparison, a component-comparability audit checks aggregation, units, and semantics.

For the calibration-first branch, AGC2 logged weather, actuator proxies, and indoor climate targets are converted into compact replay tables with blocked validation windows. Candidate GreenLight parameter sets are evaluated by replaying logged actions and comparing simulated and measured $T_{\mathrm{air}}$, relative humidity, and CO\textsubscript{2}. A candidate passes the quality gate if it improves the blocked validation objective across action-supported contexts without degrading validation temperature RMSE by more than $1^\circ$C. Only after this gate is passed are reward components recomputed on the default simulator, the candidate simulator, and the corresponding real-log rows at matched aggregation scales.

The AGC2 calibration data contain 47,809 weather rows and 23,910 action/target rows in the compact hourly contract. The raw time-block split assigns the first 80\% of each team trace to calibration and the last 20\% to validation, but the C3c search uses stricter action-supported 3-day replay windows. Within each selected 3-day window, the first half is used for candidate scoring and the second half for blocked validation. Both imputed and boiler/CO\textsubscript{2}-complete action policies are considered; windows must retain at least 75\% of expected rows and sufficient boiler/CO\textsubscript{2} support. The selected candidate is one shared GreenLight parameter set across all selected AGC2 contexts, not a separate per-team fit.

The validation objective is the sum of normalized RMSE values for air temperature, relative humidity, and CO\textsubscript{2}, using scale factors of $5^\circ$C, 20 percentage points, and 500~ppm, respectively:
\begin{equation}
J=\frac{\mathrm{RMSE}(T_{\mathrm{air}})}{5}
+\frac{\mathrm{RMSE}(RH_{\mathrm{air}})}{20}
+\frac{\mathrm{RMSE}(CO2_{\mathrm{air}})}{500}.
\end{equation}
The local search budget was 24 candidates with seed 41, initialized around a C3b warm start and filtered by the $1^\circ$C validation temperature-degradation guard. Table~\ref{tab:calibration_params} lists the searched parameter bounds and the selected C3c candidate values.

\begin{table}[htbp]
\centering
\caption{GreenLight parameters varied in the AGC2 calibration search. Bounds are the local-search ranges; the selected values are the shared candidate used for all selected AGC2 validation contexts.}
\label{tab:calibration_params}
\scriptsize
\setlength{\tabcolsep}{3pt}
\resizebox{\linewidth}{!}{%
\begin{tabular}{rllrrl}
\toprule
Index & Parameter & Role & Search range & Selected & Unit \\
\midrule
46 & aFlr & floor area & 12.5--30.0 & 14.982 & m$^2$ \\
47 & aCov & cover area & 25.0--90.0 & 72.632 & m$^2$ \\
48 & hAir & main compartment height & 3.0--7.0 & 3.000 & m \\
49 & hGh & mean greenhouse height & 4.0--9.0 & 5.211 & m \\
50 & cHecIn & cover--air heat exchange & 1.5--8.0 & 4.675 & W m$^{-2}$ K$^{-1}$ \\
56 & hVent & ventilation opening height & 0.2--2.5 & 1.101 & m \\
59 & cDgh & ventilation discharge coefficient & 0.05--0.80 & 0.110 & -- \\
60 & cLeakage & leakage coefficient & $10^{-5}$--$10^{-3}$ & $2.37\times10^{-4}$ & m$^3$ s$^{-1}$ m$^{-2}$ Pa$^{-1}$ \\
61 & cWgh & wind shelter factor & 0.005--0.120 & 0.0185 & -- \\
108 & pBoil & boiler capacity & 1000--60000 & 4403.9 & W \\
109 & phiExtCo2 & external CO\textsubscript{2} capacity & 20--3000 & 20.0 & mg s$^{-1}$ \\
172 & thetaLampMax & lamp capacity & 50--300 & 300.0 & W m$^{-2}$ \\
\bottomrule
\end{tabular}}
\end{table}

\subsection{Reproducibility and release-asset map}

The release package separates the submission-facing evidence from reusable code. The main manuscript reports the diagnostics needed for the argument, Additional file~1 preserves supplementary ablation and artifact checks, and the code repository provides reproduction scripts and component definitions for reuse on new greenhouse traces.

\section{Results}
\label{sec:results}

The decomposed reward first had to remain learnable under PPO. Subsequent checks test whether selected actuator changes occur in interpretable regimes, whether component scales are comparable, and whether AGC2 calibration changes rollout accuracy. The final checks score the same components on AGC logs, measure action-proxy tradeoffs, and distill actuator decisions into shallow rules.

\subsection{The decomposed reward remains learnable under PPO}

All nine 500k-step PPO runs clear the official GreenLight-Gym rule-based controller floor \citep{GreenLightGym} (Table~\ref{tab:500k}). The component reward matches or slightly exceeds the scalar baseline: tomato component $169.54 \pm 0.78$ vs scalar $168.82 \pm 0.99$ ($+0.72 \pm 0.45$), cucumber component $169.47 \pm 0.42$ ($+0.66 \pm 0.72$). In this training setting, the physics-prior wrapper adds interpretable component traces without reducing native simulator reward.

After facility adaptation, all six PPO runs in the AGC2-adapted simulator also cleared the rule-controller floor. Scalar-reward tomato achieved $89.526 \pm 0.663$ and component-reward tomato $89.392 \pm 0.540$. Matched-seed margins were mixed ($+0.554$, $-0.912$, $-0.042$; mean $-0.133$), so neither reward design dominates in the adapted environment. The adapted simulator supports stable policy training under both scalar and component rewards.

A matched 1M-step budget check in the adapted simulator also remained above the rule-controller floor for all seeds and gave a small directional component-reward margin (mean $+0.662$ native reward; Additional file~1). This result is retained as learnability and budget-sensitivity support rather than reward-dominance evidence.

\begin{table}[htbp]
\centering
\caption{500k-step multi-seed native reward summary (3 seeds each).}
\label{tab:500k}
\small
\begin{tabular}{p{0.34\linewidth}rrl}
\toprule
Config & Reward (mean $\pm$ std) & Margin vs scalar baseline & Status \\
\midrule
Scalar-reward tomato baseline & $168.82 \pm 0.99$ & --- & PASS \\
Component-reward tomato & $169.54 \pm 0.78$ & $+0.72 \pm 0.45$ & PASS \\
Component-reward cucumber & $169.47 \pm 0.42$ & $+0.66 \pm 0.72$ & PASS \\
\bottomrule
\end{tabular}
\end{table}

\subsection{CO\textsubscript{2} actions shift modestly toward radiation-supported regimes}

\begin{figure}[htbp]
\centering
\includegraphics[width=0.88\textwidth]{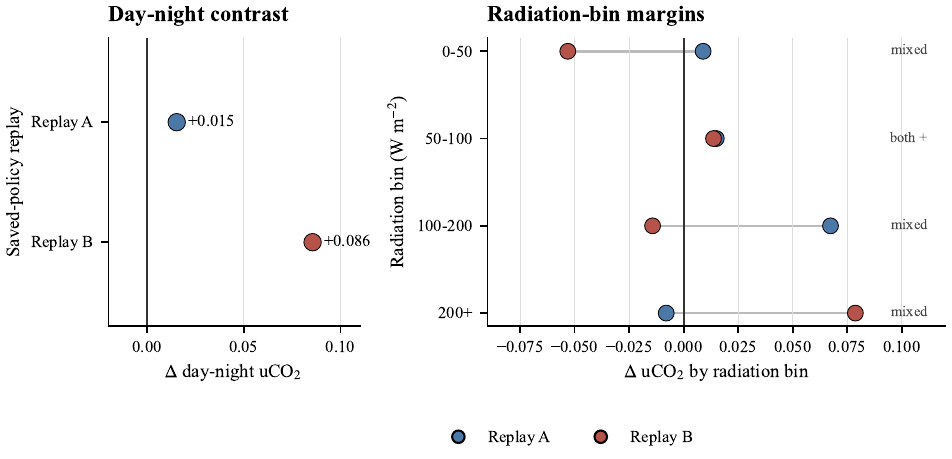}
\caption{Two-replay CO\textsubscript{2} action diagnostic. The component reward increases day/night CO\textsubscript{2} action contrast in both saved-policy replays (day $n=1680$, night $n=1200$ per replay), but radiation-bin margins are mixed across seeds. Points show component-minus-scalar normalized simulator actions, not CO\textsubscript{2} mass-flow measurements.}
\label{fig:co2_bins}
\end{figure}

Time-conditioned rollout analysis shows that the tomato component reward modestly increases day/night CO\textsubscript{2} action separation in two saved-policy replays. In replay A, the day/night contrast is $0.452$ under the component reward and $0.437$ under the scalar baseline (margin $+0.015$). In replay B, the contrast increases from $0.378$ to $0.464$ (margin $+0.086$; Fig.~\ref{fig:co2_bins}). Radiation-bin margins are less stable: the 100--200~W~m$^{-2}$ bin is positive in replay A ($+0.067$) but slightly negative in replay B ($-0.014$), even though this is the regime where the gated CO\textsubscript{2} term is physiologically motivated \citep{Mortensen1987CO2}. The two replays correspond to training seeds 42 and 456 for reproducibility. The supported statement is therefore narrow: the component reward modestly strengthens day/night CO\textsubscript{2} separation in the available replays, while radiation-bin differences remain a simulator-side diagnostic rather than a stable mechanism claim.

\subsection{Component scores require scale and semantic matching}

One-out component ablation shows that all component indices are close to zero at the normalized reward scale (largest: temperature direction $+0.0056$; see Additional file~1 for the full ranking table). The greenhouse reward operates as an interacting system in which temperature, radiation, CO\textsubscript{2}, humidity, screen, and cost terms shape actuator behavior jointly, consistent with the coupled physics of greenhouse climate control.

Before cross-domain comparison, simulator and AGC component scores were matched for aggregation scale. The simulator sums 96 timestep-level component values per day, whereas AGC logs yield one daily row. After scale matching, the apparent temperature-comfort gap fell from approximately five reward units to $0.07$--$0.10$ units. The remaining CO\textsubscript{2}-direction difference reflects a semantic gap between a timestep radiation gate and a daily proxy. Additional file~1 gives the full comparability audit table and figure.

\subsection{AGC2 calibration improves state rollout and component comparability}

\begin{figure}[htbp]
\centering
\includegraphics[width=0.92\textwidth]{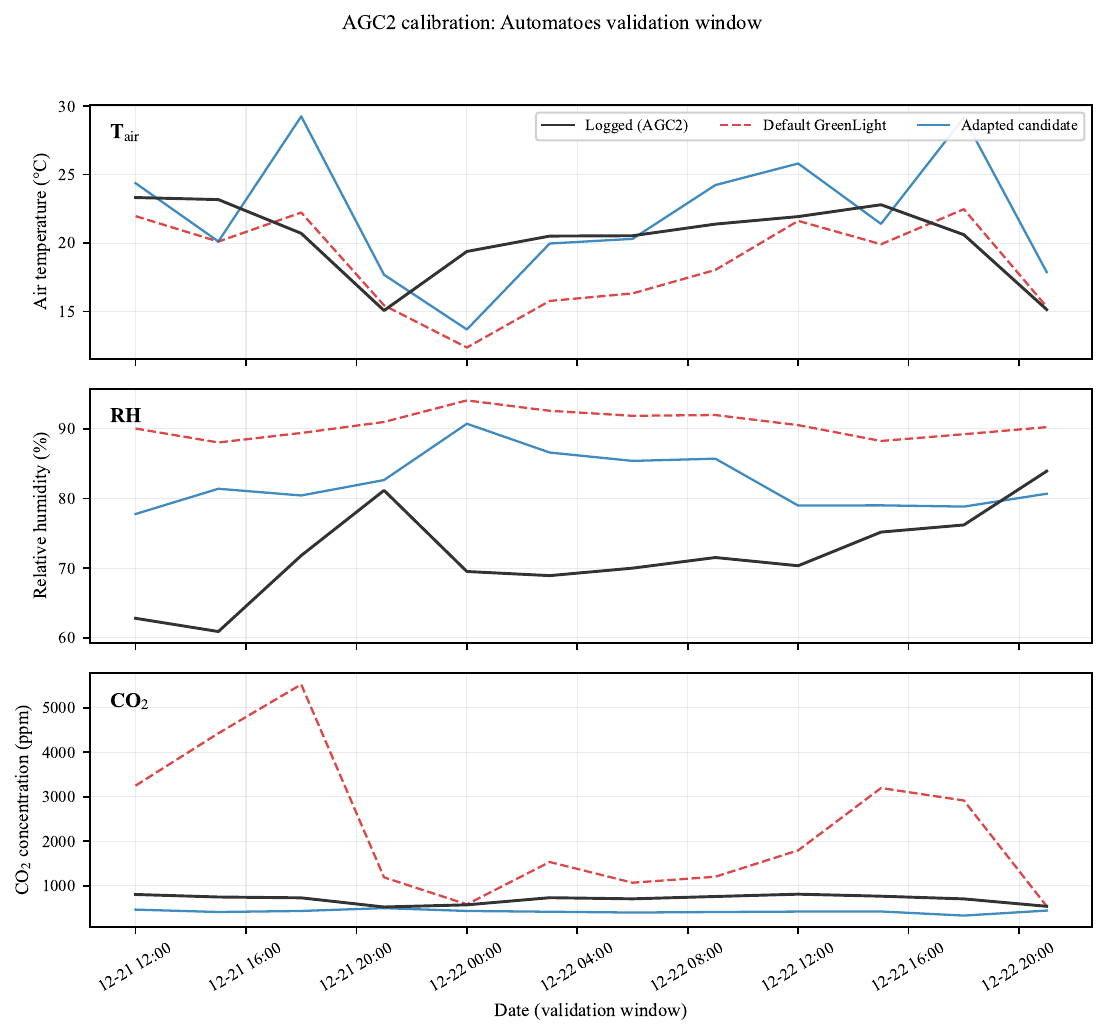}
\caption{Calibration before/after comparison on the Automatoes AGC2 validation window. Default GreenLight (dashed red) diverges substantially from logged indoor climate (solid black), especially in relative humidity and CO\textsubscript{2}. The facility-adapted candidate (blue) improves the validation objective and substantially reduces relative-humidity and CO\textsubscript{2} error, while temperature remains within the predefined no-worse-than-$1^\circ$C guard.}
\label{fig:calibration_curve}
\end{figure}

The facility-adapted GreenLight candidate improved state-rollout accuracy across four unique AGC2 team/window traces represented by seven preprocessing validation contexts (Table~\ref{tab:calibration_first_gates} and Fig.~\ref{fig:calibration_curve})\footnote{Seven contexts arise from four AGC2 teams $\times$ two preprocessing variants (impute and exclude). For three of the four teams, both variants produce identical action tables, so the seven count reflects preprocessing completeness rather than seven fully independent facility traces.}, with no context exceeding the $1^\circ$C temperature-degradation guard. Mean paired improvements over default GreenLight \citep{Katzin2020GreenLight} were $+0.184^\circ$C in temperature RMSE, $+4.274$~pp in relative-humidity RMSE, and $+562.529$~ppm in CO\textsubscript{2} RMSE across the seven preprocessing rows (positive = lower error).

\begin{table}[htbp]
\centering
\caption{AGC2 blocked-validation state-rollout deltas for the selected shared candidate. Positive values mean the candidate simulator is closer to logged AGC2 indoor climate than default GreenLight; negative temperature deltas indicate small temperature-RMSE worsening but remain within the $1^\circ$C guard.}
\label{tab:calibration_first_gates}
\scriptsize
\setlength{\tabcolsep}{3pt}
\resizebox{\linewidth}{!}{%
\begin{tabular}{llrrrrrl}
\toprule
Team & Action policy & Start day & $\Delta J$ & $\Delta T$ RMSE & $\Delta RH$ RMSE & $\Delta CO\textsubscript{2}$ RMSE & Guard \\
\midrule
AICU & impute & 1 & 0.428 & -0.026 & 4.154 & 112.795 & pass \\
Automatoes & exclude & 4 & 3.672 & -0.828 & 6.389 & 1759.021 & pass \\
Automatoes & impute & 4 & 3.672 & -0.828 & 6.389 & 1759.021 & pass \\
Digilog & exclude & 5 & 0.363 & 0.647 & 2.165 & 62.883 & pass \\
Digilog & impute & 5 & 0.363 & 0.647 & 2.165 & 62.883 & pass \\
Reference & exclude & 5 & 0.565 & 0.838 & 4.328 & 90.552 & pass \\
Reference & impute & 5 & 0.565 & 0.838 & 4.328 & 90.552 & pass \\
\midrule
Mean & --- & --- & 1.376 & 0.184 & 4.274 & 562.529 & pass \\
\bottomrule
\end{tabular}}
\end{table}

Reward-component comparability also improved after adaptation. Mean Wasserstein-1 distance between real-log and simulator component distributions decreased from 0.01677 to 0.01389 at the timestep scale and from 1.19748 to 0.97708 at the daily-sum scale. The aggregate component improved across all paired validation contexts, though the row-level improvement fraction was 62/112 (55.4\%), reflecting uneven gains across individual component channels.

\subsection{Logged AGC scores identify crop- and channel-specific alignment}

\begin{figure}[htbp]
\centering
\includegraphics[width=0.92\textwidth]{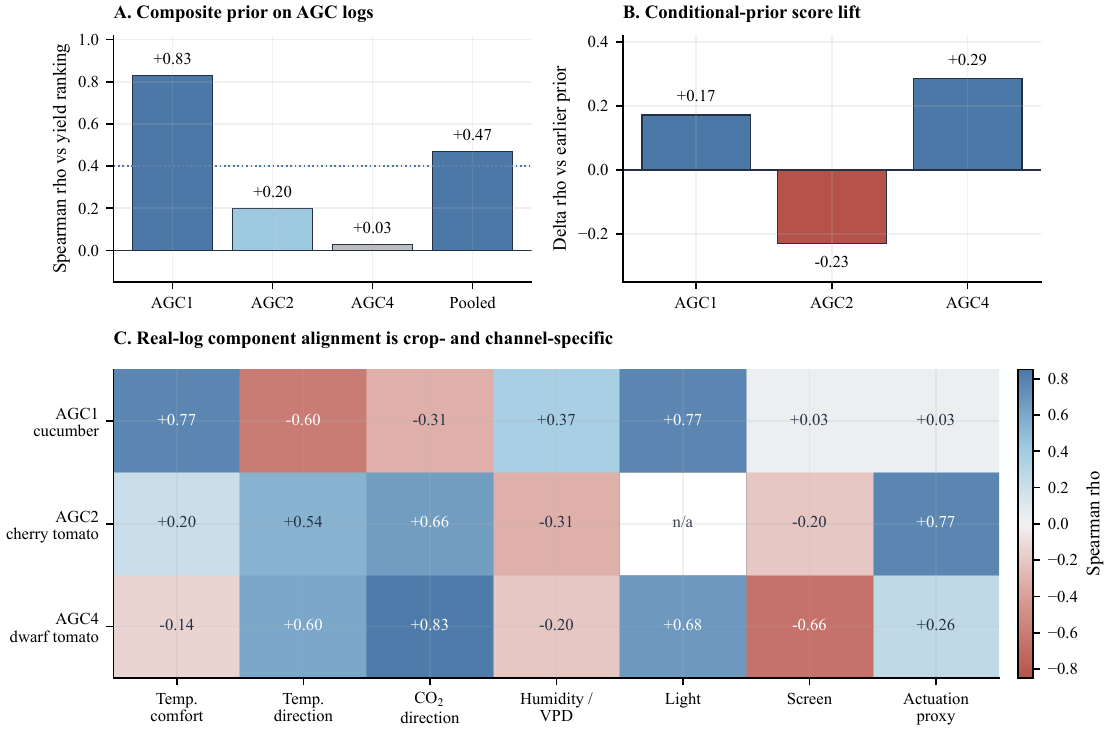}
\caption{AGC logged-data bridge for the physics-prior reward. (A) The composite prior gives a positive but underpowered AGC1 cucumber rank correlation and a weak pooled signal across AGC1/2/4 after within-dataset standardization. (B) The conditional prior improves over an earlier physics-prior variant on AGC1 and AGC4 but not AGC2. (C) Component-level correlations are crop- and channel-specific, mapping where the prior aligns with production data and where crop-specific calibration is needed.}
\label{fig:real_log_bridge}
\end{figure}

Computing the same physics-prior components on logged AGC greenhouse data \citep{Hemming2019AGC1,Hemming2020AGC2,Maree2025AGC4} links the simulator reward design to observed production records (Fig.~\ref{fig:real_log_bridge}). On AGC1 cucumber, the conditional composite prior gives a positive but underpowered team-level rank correlation ($\rho=0.829$; asymptotic $p=0.042$; exact permutation $p=0.058$; six teams), improving over an earlier physics-prior variant by $\Delta\rho=+0.172$. Across all three AGC datasets (18 team-level points), within-dataset standardized scores give a pooled correlation of $\rho=0.468$ (asymptotic $p=0.050$; Monte Carlo permutation $p=0.053$).

The alignment is crop- and channel-specific. At the composite level, AGC2 cherry tomato ($\rho=0.200$) and AGC4 dwarf tomato ($\rho=0.029$) are weak. Individual component channels are more informative: AGC2 CO\textsubscript{2} direction ($\rho=0.657$), AGC2 actuation proxy ($\rho=0.771$), and AGC4 CO\textsubscript{2} direction ($\rho=0.829$; exact permutation $p=0.058$). Bootstrap confidence intervals and permutation checks reflect the small team-level sample sizes (AGC1 composite 95\% CI [0.000, 1.000]; pooled CI $[-0.016, 0.858]$; BH-FDR $q=0.101$ for the strongest signals). These correlations are used to locate likely calibration targets, not to validate the reward as an agronomic objective.

\subsection{Actuation-proxy tradeoffs remain action-space diagnostics}

The actuation-proxy sweep reports normalized control burden as an engineering diagnostic. The component reward includes action penalties, and the available quantities are normalized simulator actions rather than audited resource meters. The sweep therefore tests whether reward gains coincide with changed action magnitudes inside the simulator. In this action space, the physics-prior reward creates favorable reward--actuation tradeoff points relative to the scalar baseline (Additional file~1, Table~S4). At actuation-proxy weight $0.5$, the component reward achieved $0.5663$ vs scalar $0.5596$, showing that the proxy can reveal whether a reward improvement is being bought through actuator shortcuts.

\subsection{Rule fidelity provides the strongest actuator-level evidence}

\begin{figure}[htbp]
\centering
\includegraphics[width=0.86\textwidth]{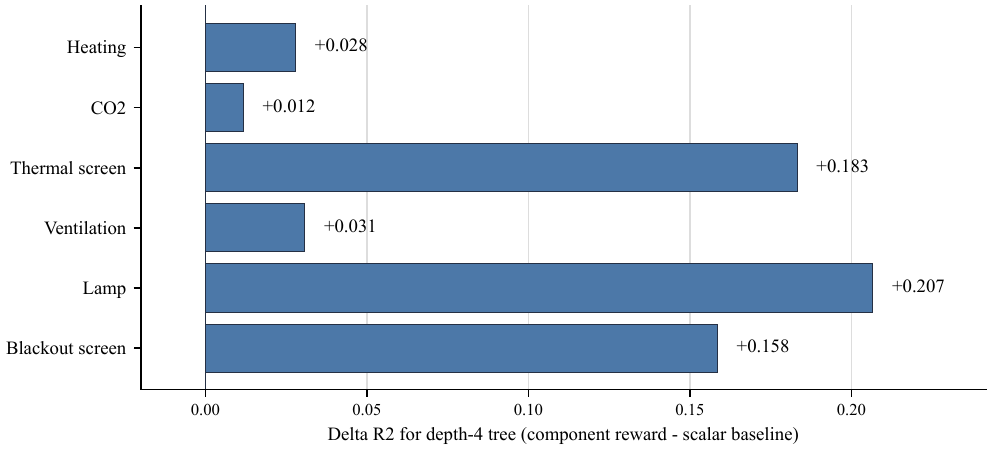}
\caption{Actuator-level change in shallow-tree rule fidelity under the physics-prior component reward relative to the scalar baseline. Positive values mean that the same action channel is more closely approximated by a depth-4 decision tree under the same rollout distribution. The largest gains occur for thermal-screen, lamp, and blackout-screen channels, indicating which actuator behaviors are most amenable to simulator-side rule inspection before facility-specific validation.}
\label{fig:rule_fidelity_delta}
\end{figure}

Rule-fidelity analysis provides the clearest behavioral evidence for the physics-prior reward. Decision-tree distillation \citep{Bastani2018VIPER} shows higher shallow-rule approximability for thermal screen, lighting, and blackout screen actions under the component reward than under the scalar baseline (Table~\ref{tab:rule_fidelity}; Fig.~\ref{fig:rule_fidelity_delta}). Thermal-screen $R^2$ rises from 0.652 to 0.835, lamp from 0.631 to 0.838, and blackout screen from 0.592 to 0.750. These are actuator channels where growers expect rules tied to radiation, temperature, and weather. In this setting, the physics-prior reward makes the main actuator decisions easier to summarize as shallow greenhouse rules.

In the facility-adapted simulator, both reward designs produce highly rule-approximable policies: mean shallow-tree $R^2$ across six actuators was 0.945 (scalar) and 0.943 (component), with random-forest $R^2$ at 0.951 and 0.956. The convergence in rule-fidelity after facility adaptation indicates that the adapted GreenLight dynamics constrain actuator behavior toward physically structured rules regardless of the training reward. Channel-level differences between reward designs are small and mixed in the adapted environment.

A depth-sensitivity replay on the same three-seed calibrated traces supports the same interpretation (Additional file~1). Using local CART surrogates with minimum leaf size 50, the mean actuator $R^2$ across six channels was already about 0.92 at depth 2 and increased smoothly to about 0.95 at depth 5 for both scalar and component rewards. The calibrated-domain rule-fidelity result is therefore not an artifact of one depth-4 tree choice; it also shows that much of the post-calibration interpretability gain comes from the adapted simulator distribution shared by both reward designs.

\begin{table}[htbp]
\centering
\caption{Depth-4 decision-tree rule fidelity for actuator channels. Higher $R^2$ means the simulator policy action is more closely approximated by a shallow rule under the same rollout distribution.}
\label{tab:rule_fidelity}
\small
\begin{tabular}{lrrll}
\toprule
Actuator & Physics-prior $R^2$ & Scalar $R^2$ & Physics-prior top feature & Scalar top feature \\
\midrule
uBoil & 0.762 & 0.734 & RHair & PAR total \\
uCO\textsubscript{2} & 0.764 & 0.753 & VPD & VPD \\
uThScr & \textbf{0.835} & 0.652 & VPD & hour sine \\
uVent & 0.688 & 0.657 & outside RH & VPD \\
uLamp & \textbf{0.838} & 0.631 & wind speed & wind speed \\
uBlScr & 0.750 & 0.592 & radiation & radiation \\
\bottomrule
\end{tabular}
\end{table}

Additional file~1 retains the full component ablation, pairwise interaction, actuation-proxy sweep, policy-distillation, and component-comparability tables.

\section{Discussion}
\label{sec:discussion}

The central contribution is a reward-component audit for greenhouse RL. Rather than relying on a single simulator return, the study follows the same named components through policy learning, AGC2 calibration, AGC log scoring, and shallow-rule actuator analysis. This gives greenhouse RL studies a practical way to report how reward is earned, which parts of the simulator need calibration, and which actuator channels are suitable for rule-based inspection.

The results separate protocol validation, simulator-side behavior, and logged-data alignment. Component rewards remained trainable, AGC2 calibration improved blocked state rollout and reward-component comparability, and default-simulator screen and lighting actions became easier to approximate with shallow rules. The AGC yield-ranking result serves as calibration-target evidence: useful for deciding what to check next, but too small to validate the reward as an agronomic objective.

\subsection{Why AGC log scoring matters}

Scoring the same reward components on AGC logs moves the reward check outside the simulator. It allows a candidate RL reward to be compared with observed crop-management records before it is adopted as a training objective. On AGC1 cucumber \citep{Hemming2019AGC1}, the crop-adjusted composite prior gives a positive but underpowered team-level rank correlation ($\rho=0.829$), with bootstrap intervals, exact permutation testing, and BH-FDR correction limiting the strength of this six-team result. AGC2 and AGC4 show crop- and channel-specific alignment. CO\textsubscript{2}-direction channels are informative in tomato logs even when the composite score is weak.

AGC1, AGC2, and AGC4 data are used here for retrospective reward-component scoring and calibration-target identification \citep{Hemming2019AGC1,Hemming2020AGC2,Maree2025AGC4}. This use is valuable because it shows which reward terms can be evaluated on public greenhouse records and which crop-specific channels need improved calibration or measurement support.

\subsection{Interpretability versus transfer}

Rule fidelity and sim-to-real transfer are different claims. The comparability check ensures that simulator and real-log component scores share aggregation, units, and semantics before cross-domain conclusions are drawn, following standard sim-to-real caution \citep{Tobin2017DomainRandomization,Salvato2021RealityGap,Muratore2022RandomizedSimReview}. The AGC2 facility-adaptation branch showed that a candidate parameter set can improve both state rollout and aggregate reward-component comparability, while the method also identifies which components still require calibration.

The convergence of rule-fidelity scores after facility adaptation ($R^2 \approx 0.94$ for both reward designs) is informative. In the default simulator, the physics-prior reward produces more rule-approximable actuator behavior. After adaptation narrows the simulator--facility gap, the adapted dynamics themselves constrain both reward designs toward similar actuator structure. This suggests that the physics-prior reward is most valuable when simulator--facility mismatch is largest.

\subsection{Implications for greenhouse RL reporting}

The results argue for reporting greenhouse RL policies with more than aggregate simulator reward. In this study, the useful reporting items are the rule-controller floor, the distinction between physical quantities and action proxies, shallow-rule summaries of actuator channels, and the ability to score the same reward components on logged greenhouse data. This reporting style connects RL studies back to the older greenhouse-control literature, where classical optimal-control studies made modelling assumptions explicit \citep{Bakker1995GreenhouseClimate,VanStraten2010OptimalControl,Vanthoor2011GreenhouseClimateModel}. The reward wrapper provides a practical link between agronomic knowledge and RL evaluation: the same named terms train a simulator policy, score production data, identify calibration targets, and summarize actuator behavior for engineering inspection.

\subsection{Interpretation limits}

Three interpretations remain outside the current evidence. The CO\textsubscript{2} diagnostic strengthens day/night separation in two saved-policy replays, but mixed radiation-bin margins do not support a stable CO\textsubscript{2} mechanism claim. Scoring AGC logs identifies crop-specific calibration targets, but the small team-level samples and permutation/FDR results do not validate the reward as an agronomic objective. The AGC2-adapted PPO runs clear the rule-controller floor under both scalar and component rewards, but their mixed 500k margins do not support component-reward dominance.

\subsection{Future calibration path}

A stronger sim-to-real study would need matched aggregation on both sides of the comparison and a simulator model that remains stable after adaptation \citep{Salvato2021RealityGap}. On the simulator side, component scores can be summarized as timestep means, daily aggregates computed after first aggregating state and action variables, and season-level summaries. On the logged-data side, action proxies can be replaced where possible by audited physical quantities such as heating energy, lamp energy, CO\textsubscript{2} dosage, and screen position. The comparison can then be repeated under weather regimes shared by simulation and data, with sensitivity analyses for start date, radiation distribution, and actuator support.

Here, GreenLight is adapted to logged AGC2 climate trajectories and the reward and policy diagnostics are repeated in the adapted candidate simulator. With richer facility data, the same loop can support a practical deployment pathway: first calibrate the process model to logged weather, actions, and climate states; then add residual correction where the process model is biased; then use the audited reward components to select candidate policies for greenhouse trials. A production-first extension would treat climate states, VPD, CO\textsubscript{2}, and light as intermediate variables for yield or biomass prediction, while keeping cost and energy terms as optional diagnostics rather than headline outcomes. A residual-hybrid simulator is a natural next direction because it can preserve process-model structure while learning facility-specific climate residuals.

\subsection{Limitations}

\begin{itemize}
  \item All policy training and rollout results are simulator-based. Prospective greenhouse trials would be needed to evaluate closed-loop operation.
  \item AGC log scoring uses small team-level samples (6--10 teams per dataset). Bootstrap confidence intervals and BH-FDR correction bound statistical power accordingly.
  \item Facility adaptation was selected and validated on AGC2 only; generalization to other facilities remains untested.
  \item Several selected AGC2 calibration parameters lie at or near the local-search bounds. The selected vector should therefore be read as a facility-adaptation coefficient set, not as a uniquely identifiable estimate of greenhouse equipment properties.
  \item Actuation-proxy tradeoffs reflect normalized simulator actions, not measured physical energy.
  \item GreenLight-Gym is a simplified greenhouse model; residual-hybrid extensions require further validation before stronger transfer claims.
\end{itemize}

\section{Conclusions}
\label{sec:conclusions}

This study establishes reward-component auditing as a reproducible intermediate step between simulator-only greenhouse RL and facility-specific validation. It shows that greenhouse RL reward design can be inspected across policy learning, calibrated rollouts, AGC log scoring, and actuator-rule distillation. The supporting evidence is that the component reward remained trainable, AGC2 facility adaptation improved state-rollout accuracy across four unique team/window traces represented by seven preprocessing validation contexts, and real-log--simulator reward-component distances decreased. The behavioral evidence is strongest in the default simulator, where thermal-screen, lamp, and blackout-screen decisions became more rule-approximable under the physics-prior reward; after facility adaptation, both scalar and component rewards became similarly rule-approximable, indicating that adapted dynamics dominate that comparison. Together, these results support reward-component auditing as a pre-deployment diagnostic layer for greenhouse RL; field-level yield gains and closed-loop controller transfer require dedicated greenhouse trials.

\section*{Data availability}

GreenLight and GreenLight-Gym are cited as external model and benchmark resources \citep{Katzin2020GreenLight,GreenLightGym}. The AGC logged data used for real-log reward-component scoring are available from the original 4TU.ResearchData releases for AGC1 \citep{Hemming2019AGC1}, AGC2 \citep{Hemming2020AGC2}, and AGC4 \citep{Maree2025AGC4}. We release the reward-audit workflow developed in this study, including reward-component definitions, the physics-prior reward wrapper, and multi-component scoring and validation code. The repository URL is \url{https://github.com/shennongwm/rl-reward-audit}.

\section*{CRediT authorship contribution statement}

Yuhui Bie: Conceptualization, methodology, software, formal analysis, visualization, writing -- original draft, writing -- review and editing. Guowei Xu: Data curation, validation, investigation, visualization, writing -- original draft, writing -- review and editing. Yaojun Wang: Conceptualization, resources, supervision, project administration, writing -- review and editing.

\section*{Declaration of competing interest}

The authors declare that they have no known competing financial interests or personal relationships that could have appeared to influence the work reported in this paper.

\section*{Funding}

This research is supported by the Joint Fund Project of the National Natural Science Foundation of China (U25B2019-1).

\section*{Acknowledgments}

The authors thank the GreenLight and GreenLight-Gym developers, and the Wageningen Autonomous Greenhouse Challenge organizers and participants for making the simulation environment and logged greenhouse datasets available.

\section*{Declaration of generative AI and AI-assisted technologies in the manuscript preparation process}

During the preparation of this work the authors used OpenAI Codex/GPT and Anthropic Claude for language polishing, code-checking assistance, and organization of revision notes. After using these tools, the authors reviewed and edited the content as needed and take full responsibility for the content of the published article.

\clearpage
\appendix
\newgeometry{margin=0.75in}

\section*{Supplementary diagnostic tables}
\addcontentsline{toc}{section}{Supplementary diagnostic tables}

This file contains the detailed diagnostic evidence behind the
preprint. The tables document reward components,
actuation proxies, rule fidelity, logged-data component scoring, and
component-comparability checks used to interpret the physics-prior reward.

\subsection*{Table S1a. Reward-component definitions used for training and AGC-log scoring}

\begin{center}
\scriptsize
\begin{tabular}{P{2.0cm}P{5.3cm}P{2.0cm}P{4.2cm}}
\toprule
Component & Simulator timestep score, inputs, and gate & Weight / scale & AGC-log daily proxy \\
\midrule
Temperature comfort & $0.1\max(0,1-|T-T_{\mathrm{sp}}|/5)$, using $T$, a radiation-derived day flag, and crop setpoints. & $w=1$; timestep reward units. & Day and night terms use crop setpoints ($24/19^\circ$C for cucumber, $21.5/18.5^\circ$C for tomato). \\
Temperature direction & If $T>T_{\mathrm{sp}}+1.5^\circ$C: $0.05u_{\mathrm{vent}}-0.03u_{\mathrm{boil}}$; if $T<T_{\mathrm{sp}}-1.5^\circ$C: $0.05u_{\mathrm{boil}}-0.03u_{\mathrm{vent}}$. & $w=1$; signed action score. & Daily score uses $\pm2^\circ$C thresholds and normalized heating/ventilation proxies. \\
\cotwo{} direction & Reward $u_{\mathrm{CO2}}$ only when day, radiation $>100$~W~m$^{-2}$, $u_{\mathrm{vent}}<0.3$, and \cotwo{} is below 800~ppm; penalize night or high-vent dosing. & $w=1$; normalized action score. & $u_{\mathrm{CO2}}=\min(\mathrm{dosage}/0.3,1)$ and $u_{\mathrm{vent}}=\mathrm{vent}/100$; crop light thresholds are 3 and 5. \\
Humidity/VPD & VPD 0.5--1.2~kPa gives $+0.04$; below 0.5 rewards ventilation and penalizes no ventilation; above 1.5 penalizes excessive ventilation. & $w=1$; signed VPD score. & Same VPD formula on daily mean $T$/RH; outside-band deficits/excesses become signed daily scores. \\
Light & Day and radiation $<100$~W~m$^{-2}$ gives $+0.02u_{\mathrm{lamp}}$; night lighting gives $-0.02u_{\mathrm{lamp}}$. & $w=1$; normalized lamp score. & Lamp proxy divided by 3 and scored on low-radiation days. \\
Screen & Day: shade reward under radiation $>400$~W~m$^{-2}$ and $T>T_{\mathrm{sp}}+2^\circ$C; shade penalty under radiation $<150$~W~m$^{-2}$. Night: insulation reward. & $w=1$; $(1-u_{\mathrm{ThScr}})$ is screen deployed. & Thermal-screen percentage normalized from 0--100\%; daily high-radiation/hot, low-radiation, and night-insulation gates. \\
Actuation proxy & $-0.03(0.3u_{\mathrm{boil}}c_{\mathrm{cold}}+0.2u_{\mathrm{CO2}}+0.15u_{\mathrm{lamp}}c_{\mathrm{dark}})$, with weather factors clipped at 0.2. & $w=1$ except proxy-weight sweep; action proxy only. & Heating proxy/10, \cotwo{} dosage/0.3, lamp proxy/3, with daily cold/dark factors. \\
\bottomrule
\end{tabular}
\end{center}

\subsection*{Table S1. Ordinary 100k-step component ablation}

\begin{center}
\scriptsize
\begin{tabular}{P{2.3cm}rrrrP{3.5cm}}
\toprule
Configuration & Native mean & Native SD & $\Delta$ vs full component reward & $\Delta$ vs scalar baseline & Interpretation \\
\midrule
Scalar-reward tomato baseline & 161.178 & 1.288 & -0.902 & 0.000 & Fixed scalar comparator for scale reference. \\
Full component-reward tomato & 162.079 & 2.411 & 0.000 & +0.902 & Learnable simulator reward, but margin is small. \\
No gated \cotwo{} term & 161.315 & 1.887 & -0.764 & +0.138 & Removing gated \cotwo{} weakens reward in this diagnostic. \\
No actuation proxy & 160.965 & 2.565 & -1.114 & -0.212 & Actuation proxy contributes to simulator reward and is tracked as normalized control burden. \\
No humidity term & 162.640 & 2.070 & +0.560 & +1.462 & Humidity has a mixed contribution, motivating component-level reporting. \\
No screen term & 161.240 & 2.484 & -0.839 & +0.063 & Conditional screen has a modest simulator contribution. \\
\bottomrule
\end{tabular}
\end{center}

\subsection*{Supplementary Figure S1. 500k-step simulator learnability}

\begin{center}
\includegraphics[width=0.66\linewidth]{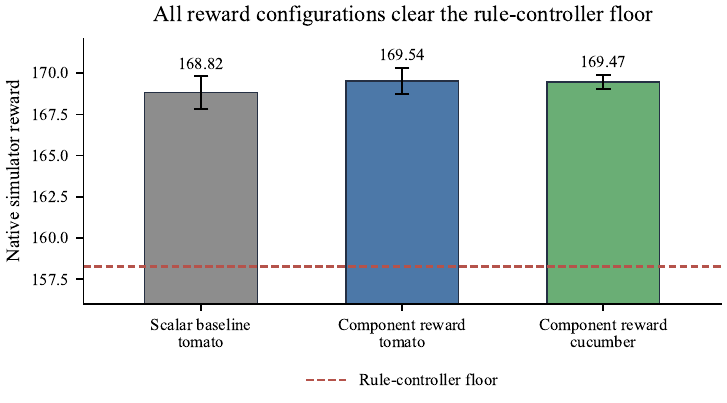}
\end{center}

\noindent All configurations clear the GreenLight-Gym rule-controller floor.
The figure documents the learnability gate used before interpreting reward
components; controller ranking is addressed separately in the main text.

\subsection*{Table S2. One-out component ranking}

\begin{center}
\small
\begin{tabular}{P{3.4cm}rP{7.3cm}}
\toprule
Component & One-out index & Interpretation \\
\midrule
Temperature direction & +0.0056 & Largest positive index, but still small at the reward scale. \\
Temperature comfort & +0.0043 & Positive and plausibly synergistic with direction terms. \\
Normalized actuation penalty & +0.0039 & Positive control-burden contribution in the simulator audit. \\
Conditional screen & +0.0029 & Modest positive simulator contribution. \\
\cotwo{} direction & +0.0027 & Modest positive simulator contribution. \\
Light & -0.0021 & Boundary/redundancy signal in this diagnostic. \\
Humidity & -0.0019 & Boundary/mismatch signal in this diagnostic. \\
\bottomrule
\end{tabular}
\end{center}

\subsection*{Table S3. Top pairwise component interactions}

\begin{center}
\small
\begin{tabular}{P{4.6cm}rP{6.0cm}}
\toprule
Interaction pair & Index & Interpretation \\
\midrule
Temperature comfort $\times$ temperature direction & +0.0099 & Temperature terms appear jointly useful in simulation. \\
Temperature direction $\times$ actuation penalty & +0.0095 & Direction and proxy-cost terms interact positively. \\
Temperature direction $\times$ conditional screen & +0.0085 & Screen behavior may be coupled to temperature direction. \\
Temperature direction $\times$ \cotwo{} direction & +0.0083 & Small positive interaction between climate correction and enrichment logic. \\
Temperature comfort $\times$ actuation penalty & +0.0082 & Comfort/cost balancing is a plausible simulator mechanism. \\
\bottomrule
\end{tabular}
\end{center}

\subsection*{Table S4. Simulator reward--actuation-proxy tradeoff}

\begin{center}
\scriptsize
\begin{tabular}{P{2.6cm}P{1.2cm}rrrr}
\toprule
Configuration & Proxy weight & Reward mean & Reward SD & Proxy mean & Proxy SD \\
\midrule
Scalar baseline & fixed & 0.5596 & 0.0045 & -0.6498 & 0.0969 \\
Component reward & 0.0 & 0.5589 & 0.0074 & -0.4799 & 0.1952 \\
Component reward & 0.3 & 0.5615 & 0.0070 & -0.5196 & 0.0507 \\
Component reward & 0.5 & 0.5663 & 0.0028 & -0.4747 & 0.0213 \\
Component reward & 0.7 & 0.5548 & 0.0048 & -0.4857 & 0.1241 \\
Component reward & 1.0 & 0.5628 & 0.0084 & -0.6094 & 0.0739 \\
Component reward & 1.5 & 0.5654 & 0.0033 & -0.5646 & 0.1798 \\
\bottomrule
\end{tabular}
\end{center}

The proxy axis is a normalized residual/action proxy. The component-reward
weight 0.5 point has the highest reward in this sweep, showing how the audit can
track whether reward changes are accompanied by changed simulator action burden.
Physical resource accounting would require matched energy, lamp, and \cotwo{}
flow measurements.

\subsection*{Table S5. Policy-distillation fidelity}

\begin{center}
\scriptsize
\begin{tabular}{P{1.8cm}rrP{2.4cm}P{2.4cm}P{3.0cm}}
\toprule
Actuator & Component reward $R^2$ & Scalar baseline $R^2$ & Component top feature & Baseline top feature & Interpretation \\
\midrule
uBoil & 0.762 & 0.734 & RHair & PAR total & Small fidelity gain. \\
uCO2 & 0.764 & 0.753 & VPD & VPD & Moderate for both; no strong simple \cotwo{} rule. \\
uThScr & 0.835 & 0.652 & VPD & hour sine & Stronger rule fidelity under component reward. \\
uVent & 0.688 & 0.657 & RH outside & VPD & Small fidelity gain. \\
uLamp & 0.838 & 0.631 & wind speed & wind speed & Stronger rule fidelity under component reward. \\
uBlScr & 0.750 & 0.592 & radiation & radiation & Stronger rule fidelity under component reward. \\
\bottomrule
\end{tabular}
\end{center}

\subsection*{Table S6. Component-comparability audit}

\begin{center}
\small
\begin{tabular}{P{2.9cm}rP{4.8cm}P{4.4cm}}
\toprule
Component & Original W1 & Audit status & Manuscript use \\
\midrule
Temperature comfort & 5.043 & Aggregation-scale contamination: timestep-summed simulator values compared with daily real-log scores. & Use only as comparability-audit motivation. \\
Temperature direction & 0.900 & Original scale not matched; daily mean rescaling reduces apparent gap. & No physical gap claim. \\
\cotwo{} direction & 0.694 & Sign/semantics differ after rescaling; magnitude is small. & State that sign convention and proxy semantics require audit. \\
Actuation proxy & 0.532 & Simulator and log paths use action proxies rather than matched resource meters. & Report as normalized control burden. \\
Humidity & 0.466 & Original scale not comparable; VPD aggregation differs. & Use after aggregation-scale matching. \\
\bottomrule
\end{tabular}
\end{center}

\subsection*{Supplementary Figure S2. Original component-gap artifact}

\begin{center}
\includegraphics[width=0.78\linewidth]{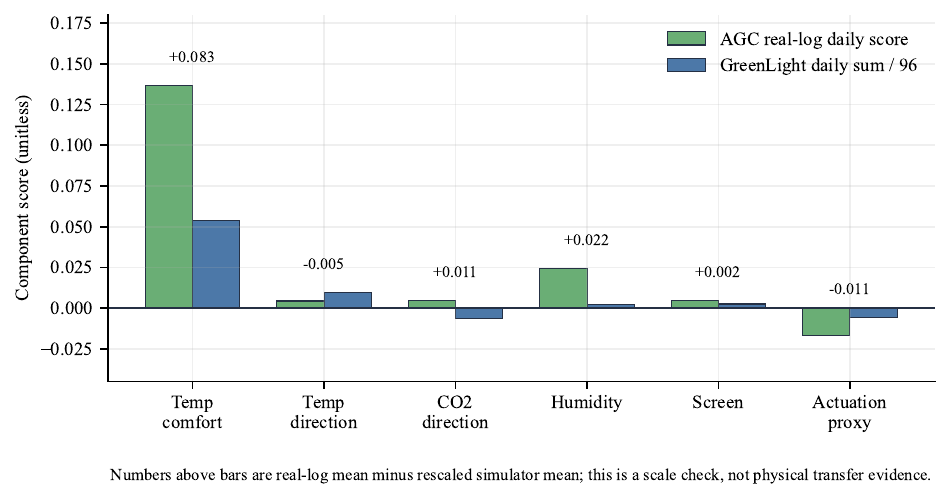}
\end{center}

\noindent This plot is retained as an audit artifact. The original
comparison mixed timestep-summed simulator values with daily aggregate
real-log scores, so it is used to document the aggregation mismatch that
motivated the component-comparability audit.

\subsection*{Scope of supplementary diagnostics}

\noindent These tables support simulator-side reward decomposition, logged-data
component scoring, and component-comparability auditing. They provide detailed
evidence for reward design and calibration planning, including which quantities
would need matched-scale resource accounting or prospective greenhouse trials
in a later validation study.

\subsection*{Table S7. Calibration-first AGC2 gates}

\begin{center}
\scriptsize
\begin{tabular}{P{3.1cm}P{3.5cm}P{3.9cm}P{3.8cm}}
\toprule
Gate & Criterion & Result & Scope \\
\midrule
C4b state rollout & Paired validation objective & Candidate \texttt{temp\_gate\_local\_005} improves 7/7 preprocessing validation rows, representing four unique team/window contexts & Facility-adapted parameter candidate \\
C4b temperature guard & Validation Tair RMSE degradation & 0/7 preprocessing validation rows degrade by more than $1^\circ$C & Allows reward-component recalibration \\
E1 timestep components & Mean real-log--simulator W1 & Default 0.01677; candidate 0.01389 & Matched-scale component comparability \\
E1 daily components & Mean real-log--simulator W1 & Default 1.19748; candidate 0.97708 & Matched-scale component comparability \\
E1 row support & Component rows improved & 62/112 rows improve; total improves in all paired contexts & Cautious PPO preparation only \\
\bottomrule
\end{tabular}
\end{center}

\subsection*{Supplementary Figure S3. AGC2 calibration deltas across unique contexts}

\begin{center}
\vspace{0.4em}
\includegraphics[width=0.96\linewidth]{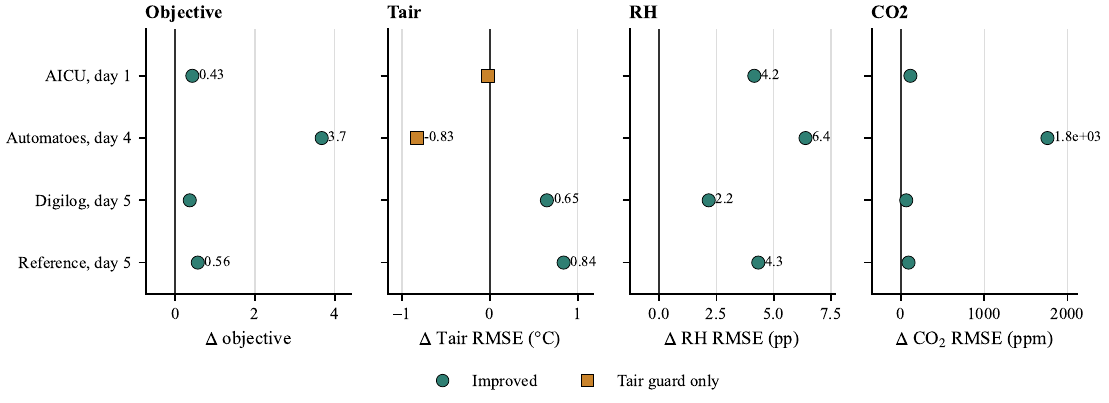}
\end{center}

\noindent Positive deltas indicate that the facility-adapted GreenLight
candidate is closer to logged AGC2 indoor climate than the default
GreenLight simulator. The four rows are unique team/window contexts;
preprocessing variants that produced identical action tables are not counted
as independent facility traces. Objective, relative-humidity, and
\cotwo{} RMSE improve in all four contexts, while temperature RMSE improves
in two contexts and remains within the no-worse-than-$1^\circ$C guard in the
two boundary contexts. This figure is included as anti-cherry-picking
support for the AGC2 calibration gate, not as evidence of generalized
facility adaptation.

\subsection*{Table S8. Calibrated-candidate PPO learnability}

\begin{center}
\small
\begin{tabular}{P{3.6cm}rrrP{4.5cm}}
\toprule
Configuration & Seeds & Final native reward & Delta vs rule floor & Interpretation \\
\midrule
Candidate scalar-reward tomato & 3 & $89.526 \pm 0.663$ & $+22.690$ & Clears rule floor in all seeds. \\
Candidate component-reward tomato & 3 & $89.392 \pm 0.540$ & $+22.556$ & Clears rule floor in all seeds. \\
Matched component--scalar margin & 3 & $-0.133 \pm 0.737$ & --- & Mixed ordering; no dominance claim. \\
\bottomrule
\end{tabular}
\end{center}

\subsection*{Table S9. 1M-step PPO budget-sensitivity}

\begin{center}
\scriptsize
\resizebox{\textwidth}{!}{%
\begin{tabular}{rrrrrrP{2.9cm}}
\toprule
Seed & Scalar final native & Component final native & Component--scalar & Scalar $\Delta$ vs floor & Component $\Delta$ vs floor & Interpretation \\
\midrule
42 & 88.096 & 89.387 & +1.291 & +21.260 & +22.551 & Both clear floor. \\
123 & 89.383 & 89.671 & +0.288 & +22.547 & +22.835 & Both clear floor. \\
456 & 88.420 & 88.827 & +0.406 & +21.584 & +21.991 & Both clear floor. \\
\midrule
Mean & 88.633 & 89.295 & +0.662 & +21.797 & +22.459 & Directional, small margin. \\
\bottomrule
\end{tabular}%
}
\end{center}

\noindent This matched-seed 1M-step check reports learnability and
budget-sensitivity in the AGC2-adapted GreenLight-Gym environment. All
scalar and component PPO runs clear the adapted rule-controller floor
without early termination. The component-reward runs are directionally
ahead at 1M steps, but the margin is small and simulator-only; stronger
claims about reward dominance, policy transfer, deployment readiness, or
measured resource savings require prospective evidence beyond this
table. The compact source table is provided as
\texttt{e2\_1m\_budget\_sensitivity.csv}.

\subsection*{Table S10. Calibrated-candidate rule-fidelity summary}

\begin{center}
\small
\begin{tabular}{P{3.9cm}rrP{5.5cm}}
\toprule
Metric & Scalar reward & Component reward & Interpretation \\
\midrule
Mean shallow-tree $R^2$ across six actuators & 0.945 & 0.943 & Near-tied distillability. \\
Mean random-forest $R^2$ across six actuators & 0.951 & 0.956 & Slight component advantage under nonlinear surrogate. \\
Finite trace rows per run & 2592 & 2592 & All six formal traces complete. \\
Terminations / NaN / OOM / ODE collapse & 0 & 0 & Stable simulator-side diagnostic run. \\
\bottomrule
\end{tabular}
\end{center}

\subsection*{Table S11. Calibrated rule-fidelity depth sensitivity}

\begin{center}
\small
\begin{tabular}{P{3.7cm}rrrr}
\toprule
Configuration & Tree depth & Runs & Mean actuator $R^2$ & Mean actuator MAE \\
\midrule
Component reward & 2 & 3 & 0.920 & 0.0304 \\
Component reward & 3 & 3 & 0.934 & 0.0240 \\
Component reward & 4 & 3 & 0.942 & 0.0213 \\
Component reward & 5 & 3 & 0.949 & 0.0191 \\
Scalar reward & 2 & 3 & 0.921 & 0.0264 \\
Scalar reward & 3 & 3 & 0.933 & 0.0200 \\
Scalar reward & 4 & 3 & 0.941 & 0.0172 \\
Scalar reward & 5 & 3 & 0.946 & 0.0151 \\
\bottomrule
\end{tabular}
\end{center}

\noindent Depth sensitivity was recomputed from the same six calibrated
trace samples used for Table S10, using local CART surrogates with
minimum leaf size 50 and the same feature and actuator target columns.
It is included only as a stability check for the tree-depth choice:
both reward designs remain highly rule-approximable from depth 2 to 5,
and component--scalar differences remain small and channel-mixed after
facility adaptation. The compact source tables are provided alongside
this supplement as the depth-sensitivity summary, delta, macro, and raw
CSV files.

\subsection*{Table S12. CO\textsubscript{2} time-conditioned seed replication}

\begin{center}
\scriptsize
\begin{tabular}{rrrrrP{4.8cm}}
\toprule
Train seed & Scalar day--night & Component day--night & Margin & Gate & Interpretation \\
\midrule
42 & 0.437 & 0.452 & +0.015 & PASS & Same seed shown in Fig.~2; medium-radiation bin has the largest positive margin. \\
456 & 0.378 & 0.464 & +0.086 & PASS & Replicates stronger day/night separation; radiation-bin margins are mixed. \\
\bottomrule
\end{tabular}
\end{center}

\begin{center}
\scriptsize
\begin{tabular}{rP{2.8cm}rrrr}
\toprule
Train seed & Radiation bin & n & Scalar uCO2 & Component uCO2 & Margin \\
\midrule
42 & 0--50 & 1457 & 0.227 & 0.236 & +0.009 \\
42 & 50--100 & 296 & 0.606 & 0.621 & +0.015 \\
42 & 100--200 & 328 & 0.711 & 0.779 & +0.067 \\
42 & 200+ & 799 & 0.966 & 0.958 & -0.008 \\
456 & 0--50 & 1457 & 0.303 & 0.250 & -0.053 \\
456 & 50--100 & 296 & 0.649 & 0.663 & +0.014 \\
456 & 100--200 & 328 & 0.762 & 0.747 & -0.014 \\
456 & 200+ & 799 & 0.913 & 0.992 & +0.079 \\
\bottomrule
\end{tabular}
\end{center}

\noindent The time-conditioned CO\textsubscript{2} replay supports a narrow
simulator-side statement: the tomato component reward increased the
day/night CO\textsubscript{2} contrast in the two available saved-policy
time-conditioned replays. It does not support a three-seed radiation-bin
mechanism claim. Seed 123 is available in the start-day action-bin audit,
but that audit explicitly bins only by evaluation start day and cannot
support daylight-gated CO\textsubscript{2} timing. Compact source files are
provided alongside this supplement as CO\textsubscript{2} seed-replication
summary and radiation-bin CSV files.

\subsection*{Table S13. Real-log bridge permutation audit}

\begin{center}
\scriptsize
\begin{tabular}{P{3.8cm}rrrrP{3.8cm}}
\toprule
Row & $n$ & Spearman $\rho$ & Asymptotic $p$ & Permutation $p$ & Method \\
\midrule
AGC1 composite & 6 & 0.829 & 0.042 & 0.058 & Exact two-sided, 720 permutations \\
AGC2 \cotwo{} direction & 6 & 0.657 & 0.156 & 0.175 & Exact two-sided, 720 permutations \\
AGC2 actuation proxy & 6 & 0.771 & 0.072 & 0.103 & Exact two-sided, 720 permutations \\
AGC4 \cotwo{} direction & 6 & 0.829 & 0.042 & 0.058 & Exact two-sided, 720 permutations \\
Pooled z-scored composite & 18 & 0.467 & 0.050 & 0.053 & Monte Carlo two-sided, 100,000 permutations \\
\bottomrule
\end{tabular}
\end{center}

\noindent The exact permutation check was added because the AGC bridge uses
small team-level samples. It confirms that the strongest logged-data rows are
best treated as directional calibration signals. These rows support
calibration-target discovery and component-prior auditing; prospective
greenhouse trials would be the next step for confirmatory reward validation
and policy-transfer assessment. The complete permutation table is included in
the package CSV and JSON audit files.

\subsection*{Table S14. AGC2 selected parameters at effective search bounds}

\begin{center}
\scriptsize
\begin{tabular}{P{2.2cm}rrrrP{5.4cm}}
\toprule
Parameter & Index & Selected & Effective low & Effective high & Interpretation \\
\midrule
hAir & 48 & 3.000 & 3.000 & 7.000 & Selected at the lower search bound. \\
phiExtCo2 & 109 & 20.000 & 20.000 & 3000.000 & Selected at the lower search bound. \\
thetaLampMax & 172 & 300.000 & 50.000 & 300.000 & Selected at the upper search bound. \\
cDgh & 59 & 0.110 & 0.050 & 0.800 & Near the lower search edge. \\
pBoil & 108 & 4403.864 & 1000.000 & 60000.000 & Near the lower search edge. \\
\bottomrule
\end{tabular}
\end{center}

\noindent Boundary-selected values are reported as transparency evidence for
the AGC2 facility-adaptation search. They should be interpreted as a compact
coefficient set that improves replay fidelity under the stated gate, not as
uniquely identifiable greenhouse equipment estimates. The full compact source
table is provided in the SAT experiment-closure parameter-boundary CSV.

\noindent Tables S7--S14 support the calibration-first manuscript update. They
establish the supplement-level evidence for simulator adaptation,
component-comparability, learnability, rule-fidelity, and time-conditioned
action diagnostics, and they define the measurements needed for later
prospective greenhouse-trial evaluation.

\restoregeometry

\bibliographystyle{elsarticle-num}
\bibliography{references}

\end{document}